\documentclass[letterpaper, 10 pt, conference]{ieeeconf}  

\IEEEoverridecommandlockouts                              

\usepackage{graphicx} 
\usepackage{stfloats}
\usepackage{amsmath} 
\usepackage{outlines}
\usepackage{array}
\usepackage{tabu} 
\usepackage[table]{xcolor}
\usepackage{adjustbox}

\title{\LARGE \bf Holdable Haptic Device for 4-DOF Motion Guidance}

\author{Julie M. Walker$^{1}$, Nabil Zemiti$^{2}$, Philippe Poignet$^{2}$, and Allison M. Okamura$^{1}$
\thanks{*This work was supported by a National Science Foundation Graduate Research Fellowhsip, a Chateaubriand Fellowship, National Science Foundation Grant 1812966, a Stanford Graduate Fellowship, and the Human-centered AI seed grant program from the Stanford AI Lab, Stanford School of Medicine and Stanford Graduate School of Business.}
\thanks{$^{1}$J.M. Walker and A.M. Okamura are with the Department of Mechanical Engineering,
        Stanford University, Stanford, CA 94305, USA
        {\tt\small juliewalker, aokamura@stanford.edu}}%
\thanks{$^{2}$N. Zemiti and P. Poignet are with the Laboratory of Informatics, Robotics and Microelectronics of Montpellier (LIRMM), University of Montpellier, CNRS,
        Montpellier, France
        {\tt\small nabil.zemiti, philippe.poignet@lirmm.fr}}%
}

\begin{document}

\maketitle
\thispagestyle{empty}
\pagestyle{empty}

\begin{abstract}

Hand-held haptic devices can allow for greater freedom of motion and larger workspaces than traditional grounded haptic devices. They can also provide more compelling haptic sensations to the users' fingertips than many wearable haptic devices because reaction forces can be distributed over a larger area of skin far away from the stimulation site. This paper presents a hand-held kinesthetic gripper that provides guidance cues in four degrees of freedom (DOF). \mbox{2-DOF} tangential forces on the thumb and index finger combine to create cues to translate or rotate the hand. We demonstrate the device's capabilities in a three-part user study. First, users moved their hands in response to haptic cues before receiving instruction or training. Then, they trained on cues in eight directions in a forced-choice task. Finally, they repeated the first part, now knowing what each cue intended to convey. Users were able to discriminate each cue over 90\% of the time. Users moved correctly in response to the guidance cues both before and after the training and indicated that the cues were easy to follow. The results show promise for holdable kinesthetic devices in haptic feedback and guidance for applications such as virtual reality, medical training, and teleoperation.

\end{abstract}

\section{Introduction}
Humans regularly receive guidance from electronic devices through visual or audio cues. A smart phone can give verbal instructions to drivers or show arrows on its screen. Augmented reality headsets can provide instructions and show trajectories to users as they complete assembly or manipulation tasks~\cite{Wang2016}. The sense of touch is a rich platform for this kind of information. Touch cues can encode direction, magnitude, and timing. They can be discreet, whereas audio cues may be disruptive. 
Additionally, they can be easier to process than visual or auditory cues when attention is divided. 
Touch guidance is central to learning in the physical world: a tennis instructor may adjust a novice's racket, and a guitar teacher may move a student's fingers. Especially for tasks that have temporal components, guidance through the sense of touch can be beneficial in training~\cite{Sigrist2013}. 
In teleoperation tasks, such as minimally invasive robotic surgery or remotely operating a robot to complete tasks around a home, haptic feedback and guidance may be able to improve performance in ways that visual or audio information cannot. 

\begin{figure}
  \centering
  \includegraphics[width=3.5in]{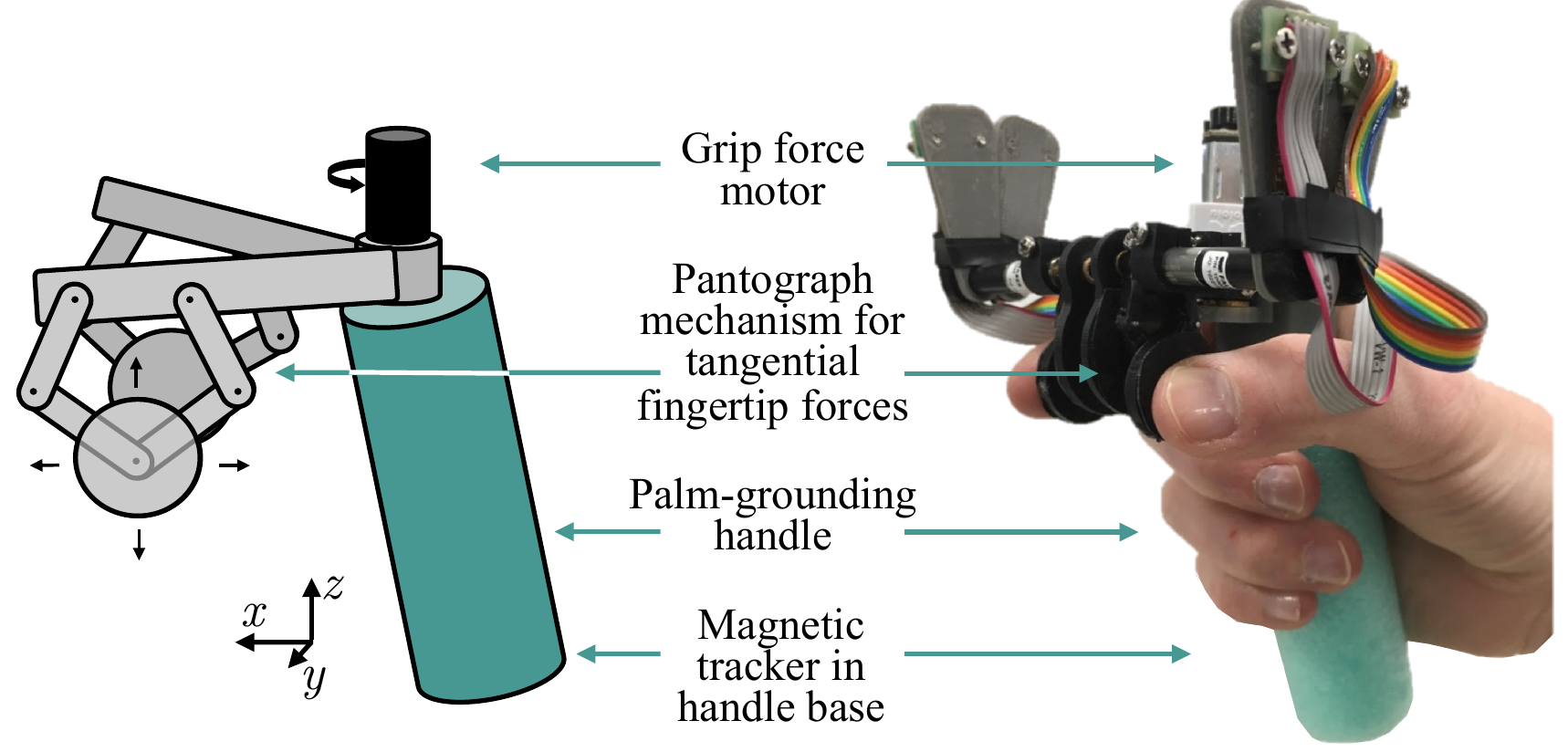}
  \caption{In this holdable haptic device, two pantograph mechanisms provide tangential displacements to a user's fingertips. The reaction forces are distributed at the handle, and thus are not noticeable. In this way, 4-DOF guidance can be generated: 2-DOF translation and 2-DOF rotation. A motor at the hinge between the arms allows the fingers to open and close and provides the potential for grip force feedback.}
  \label{fig:device}
\end{figure}
Traditional kinesthetic devices can provide clear forces or torques to move a user's hand or resist motion~\cite{Culbertson2018}. Unfortunately, they the end-effector must be physically grounded through rigid links. Larger workspaces require larger, more expensive devices. Holdable or wearable devices can be mobile and allow more freedom of motion, but it is more difficult to provide compelling directional information because they lack grounding to a world coordinate frame. 

Although they cannot provide net forces or torques, exoskeleton designs or holdable devices can generate guidance forces on the fingertips because reaction forces can be distributed across a larger area such as the user's palm. Holdable devices are easy to use because they can be picked up and put down without needing to attach straps or elastics to the fingers. They may also integrate more naturally with virtual reality systems, as they are more similar to existing controllers. So far, most holdable devices for virtual reality or teleoperation have focused on applying normal forces to the users' fingerpads or recreating textures~\cite{Benko2016,Whitmire2018}. Their use for motion guidance beyond navigation has not been fully explored. 

In this paper, we present a holdable device capable of providing four-degree-of-freedom (DOF) motion guidance (Fig.~\ref{fig:device}). Attached to handle are two 2-DOF pantograph mechanisms that displace the user's thumb and index finger to elicit pulling sensations in various directions. We show that it is able to prompt users to move their hand up/down and forward/backward as well as to twist (flex/extend) and tilt (pronate/supinate) their wrists. 

\subsection{Related Work}
\subsubsection{Wearable Cutaneous Devices}
Several wearable devices applying cutaneous cues to the fingertips or hand have shown promise for virtual object interaction and teleoperation~\cite{Schorr2017,Chinello2017}. Although they provide directional information to the fingerpad, reaction forces felt on the back of the finger can make guidance cues confusing. These devices must be fixed carefully to the hand and take some time to don and doff. Wearable vibrotactile guidance can be easy to implement, but requires the user to learn a vibration pattern mapping and interpret each cue before acting~\cite{Kapur2010,Gunther2018}. 
Asymmetric vibrations can generate intuitive direction cues for simple guidance, but the actuators must be attached very precisely to the user's hand~\cite{Culbertson2017}. 

\subsubsection{Holdable Guidance Devices}
Net forces can be generated by holdable haptic devices through air jets~\cite{Romano2009} or propellers~\cite{Heo2018}. Some researchers have had success with net-zero guidance cues for navigation through holdable haptic ``compasses." They use shape changing~\cite{Spiers2017}, gyroscopic effects~\cite{Gosselin2016}, asymmetric vibration pulses~\cite{Amemiya2009}, or weight shifting~\cite{Hemmert2010}. Higher DOF guidance can be produced by reorienting the gyroscopic or vibration pulses~\cite{Amemiya2009,Walker2017}, but these devices must use an asymmetric pulsing pattern that can be unpleasant in extended use. 

Our holdable device provides compelling guidance cues in 4-DOF by displacing both the thumb and index finger in \mbox{2-DOF}. In a user study, we demonstrate that guidance in each DOF is easy to discriminate and intuitive to follow.

\section{Device Design and Implementation}
\begin{figure}
  \centering
  \includegraphics[width=3.3in]{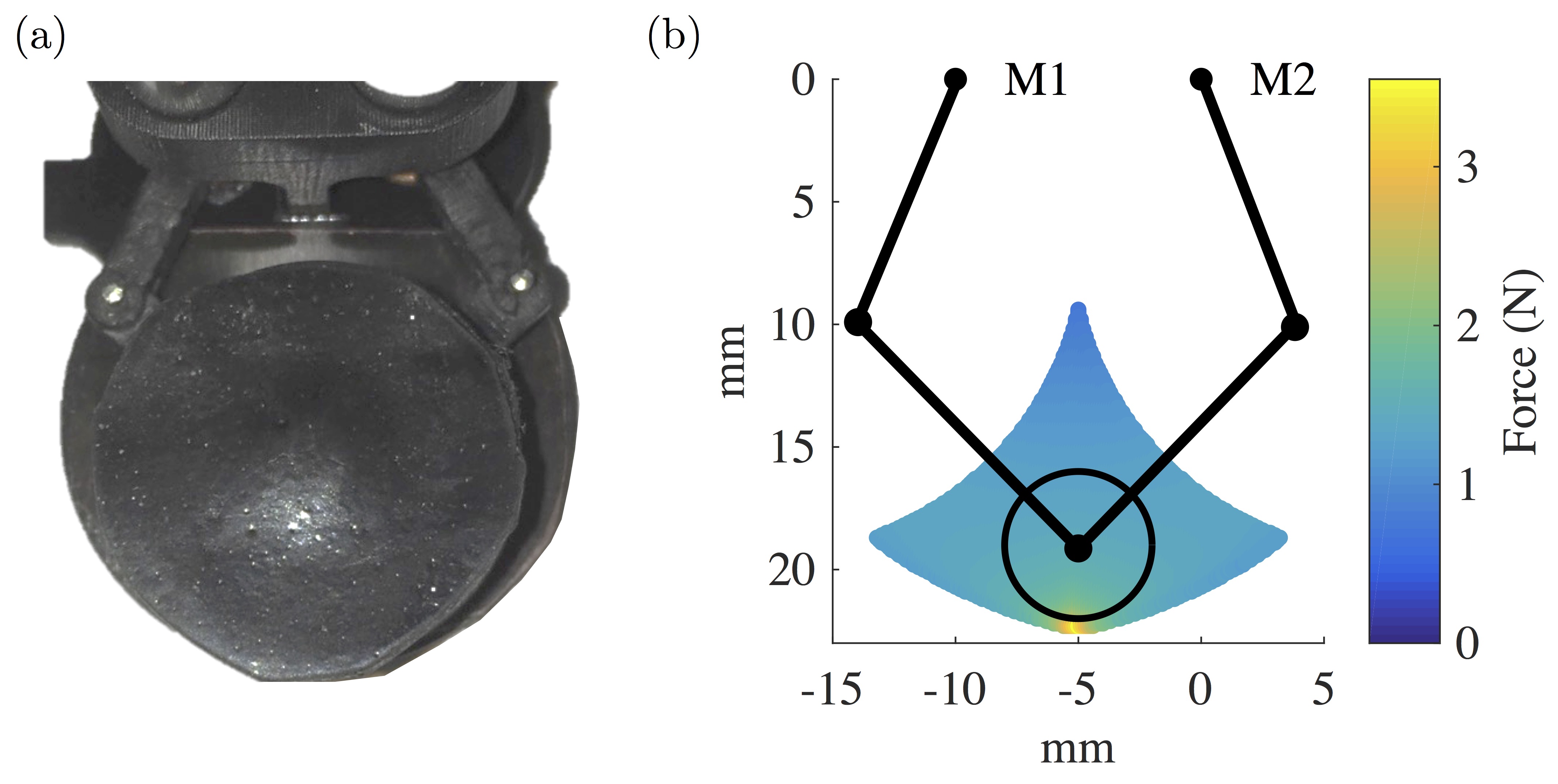}
  \caption{(a) Pantograph mechanism. (b) Pantograph workspace and isotropic force output: the maximum force that can be applied in all directions from each point in the workspace. The circle indicates the region of the workspace where the cues took place during the experiments in Section III.}
  \label{fig:isotropicForce}
\end{figure}
The device was inspired by robotic surgery simulators, which allow for movement in 6-DOF as well as gripping. We aimed to design a holdable alternative to these grounded joysticks that could include haptic guidance. Therefore, the design requirements were as follows:
\begin{itemize}
    \item allow for two-fingered gripping interactions in virtual environments or during teleoperation
    \item give directional haptic cues 
    \item allow for free space motion through a large workspace
\end{itemize}

Skin at the fingertips is highly sensitive, and fingertips are more sensitive to tangential forces than normal forces~\cite{Biggs2002}. For these reasons, we focused on applying cues tangentially to the thumb and index finger while holding a gripper. This includes up/down and distal/proximal (forward/backward) translations for each finger. Additionally, we can take advantage of the fact that there are two fingers involved in gripping and a natural center of motion at the point being pinched by the gripper. By applying tangential cues in opposite directions, the device can generate a torque sensation, cuing the user to rotate in the extension/flextion directions (twist) or the pronation/supination directions (tilt).
\subsection{Design}
As shown in Figure~\ref{fig:device}, pantograph mechanisms are attached to two arms extending from a handle. 
Pantograph 5-bar linkages are common mechanisms in kinesthetic interfaces~\cite{Campion2005}. 
In our device, each pantograph has 10 mm long upper links and 13 mm long lower links (Fig.~\ref{fig:isotropicForce}(a)). The links and all mountings are 3D printed rigid polyurethane. Metal pins connect each joint, and a circular pad rotates freely on the bottom pin. Faulhaber coreless micro DC motors (64:1 gear ratio) with optical encoders (50 counts/rev) power each upper joint. The forward and inverse kinematics equations are given by~\cite{Campion2005}, and link lengths were selected based on guidance from~\cite{Hayward1994}. The isotropic force output of the end-effector is given in Fig.~\ref{fig:isotropicForce}(b). The end-effector stayed within the 3mm radius circle during experiments (described in Section III) to maintain high quality forces in each direction. A nylon backing behind each end-effector slides against a smooth support connected to each arm. Rather than fix the end-effectors to the users' fingers with straps, we opted to cover the pads with a high-friction rubber material. As the end-effector moves, this creates a combination of skin stretch and kinesthetic feedback, displacing both the fingerpad flesh and the entire fingertip. The reaction forces simultaneously applied to the palm are not noticeable, and the user feels almost as if an object held in a precision grip is being pulled or rotated by an external force. 

The arms are attached by intermeshing gears to a geared DC motor (Pololu 50:1 micro metal gear motor) with magnetic encoder. The gears allow the arms to open and close symmetrically without the handle rotating. The motor can be used to apply a constant force holding the pads against the users' fingers as they open or to provide grip force feedback if used to interact with virtual objects. Because the studies presented in this paper focus on guidance, this motor was not used. 
The handle has a stiff core and a foam exterior. The handle adjusts to accommodate different length fingers. The entire device weighs 76.0 g.

\subsection{System}
 An Ascension TrakStar 6-DOF magnetic tracking system records the haptic device's position and orientation at 80~Hz. A sensor is mounted in the bottom of the handle. This location was chosen to keep the motors from affecting the tracking quality, which was checked using Ascension's proprietary software. The sensor's position and orientation is transformed to a point centered on the handle at the height of the pantograph end-effector. 
A Sensoray 826 PCI card controls the system. Linear current amplifiers (LM675T, 0.1 A/V gain) and a 13V external power supply (Mouser) power the motors. The motor positions $\theta$ are current-controlled using a proportional-derivative controller:
\begin{equation}
i = \frac{k_p\theta_{\text{err}}+k_d\dot{\theta}_\text{err}}{64k_t}\text{,}
\end{equation}

\noindent where the gains $k_p = 5.5$ Nm/rad and $k_d = 0.004$ Nm-s/rad, and the torque constant $k_t = 0.00196$ Nm/A. The system is run at 830 Hz in an application written using Visual C++. 

\begin{figure*}
 \centering
 \includegraphics[width=\textwidth]{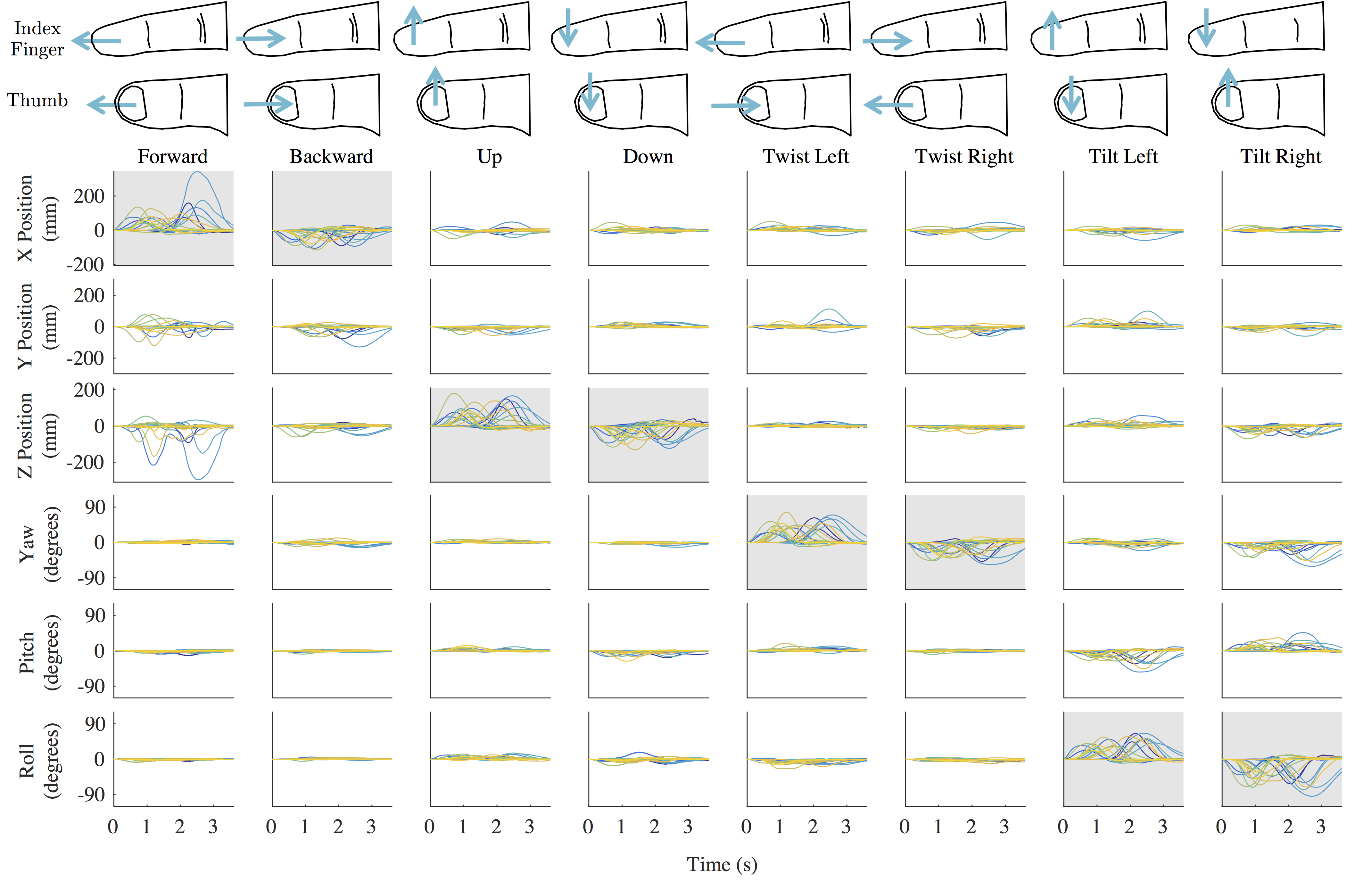}
 \caption{Each subject's mean trajectory for each cue, including Part 1 and Part 3 together. Trajectories are separated by DOF. The intended motion direction is highlighted for each cue. Each subject's data is shown with a different color line.}
    \label{fig:allMovements}
\end{figure*}

\section{User Study}

\subsection{General Methods}
In a user study we investigated two hypotheses:
\begin{enumerate}
    \item 4-DOF cues can be easily discriminated by users
    \item the guidance cues are intuitive (don't need to be explained to users in advance)
\end{enumerate}

The user study had three parts. First, users moved their hands in response to eight guidance cues (positive and negative in each DOF), without an explanation of each cue's intent. Then, users were shown what each cue was instructing them to do. They completed a forced-choice task identifying each of the cues. Finally, they repeated the first part, moving their hands in response to the cues.

20 right-handed users aged 22 to 42 participated in the experiment. All participants completed all three parts of the study.
Seven were female, and thirteen were male. 12 of the users had experience with haptic devices beyond just a few demonstrations. The protocol was approved by the Stanford University Institutional Review Board, and the subjects gave informed consent. 
In all parts of the study, users stood in front of a computer, wearing headphones to mask any noises. Users held the device with their right hand and aligned their forearm with a marking on the table at a 45$^\circ$ angle in front of their body. They were instructed to hold their wrist in a neutral position that allowed for rotation in all directions. The hand and the haptic device were hidden by a handkerchief. 

 Eight direction cues were included in the study, which we call forward (distal), backward (proximal), up, down, twist left (extension), twist right (flexion), tilt left (pronation), and tilt right (supination). We expected to see displacement in x for forward/backward cues, displacement in z for up/down cues, yaw rotation (about the z axis) for twisting cues, and roll rotation (about the x axis) for tilting cues. We expected to see minimal y and pitch displacement, which are not included. For translation cues, the pantograph end-effectors move in the same direction either up, down, forward, or backward. For rotation cues, the end-effectors move in opposite directions. The reference frame in Fig.~\ref{fig:device} illustrates the axis conventions used. 
 Fig.~\ref{fig:allMovements} shows the directions of the end-effector motion for each cue. All cues had the same magnitude and speed. 
 The end-effectors each moved 3 mm from the center position over 0.2~s, paused at the peak for 0.6~s, and returned to the center position more slowly, over 0.5~s (shown in Fig.~\ref{fig:delaydistributions}(a)). Although the end-effectors displace the fingertip in addition to the skin, part of the sensation is tangential skin-deformation. 1 mm skin deformations in four cardinal directions at 4 mm/s were discriminated with 100\% success in~\cite{Gleeson2010}, so we selected higher displacements and speeds to ensure they would be easily felt. 

\subsection{Part 1: Movement Experiment Before Training}
In the first part of the study, we asked users to move their hands in response to the haptic guidance cues. Users were not told that there were eight different cues or what they were supposed to convey. We hoped to understand whether the device's cues are intuitive. After completing 24 practice trials to get used to the sensation and the interface, users completed 80 experimental trials, experiencing each cue 10 times in a randomized order. They started each trial by pressing the space bar could repeat cues if desired. We recorded the position and orientation of the user's hand throughout.

\subsection{Part 2: Training and Forced Choice Experiment}
Before starting the second part of the experiment, the experimenter demonstrated to the users what motion each cue intends to convey. Then users felt cues in a pseudorandom order and identified them from the eight choices. A diagram of the cues similar to Fig.~\ref{fig:allMovements} was shown during the trials. They completed one practice trial of each cue followed by 24 randomized experimental trials. Again, users began each trial by pressing the space bar, and could repeat the cue if desired. They selected one direction from eight labeled keys. If incorrect, the correct response was shown on the screen to help the users learn.

\subsection{Part 3: Movement Experiment After Training}
In the third part of the experiment, users repeated the movement experiment from the initial part of the experiment. Now, they knew the eight possible cues and had practiced discriminating them in the second part of the experiment. 

\subsection{Analyses}
We analyzed the effects of cue, set, and subject on the motion directions and delay times for Parts 1 and 3. We analyzed the ease of discriminating cues in Part 2. MATLAB's Statistics and Machine Learning Toolbox was used for all analyses.

\emph{Motion Delay:} Delay time was identified as the time from the start of a cue to the first peak of acceleration in any DOF. 
We fit a linear regression model predicting delay from set number, experience with haptics (self-rated by users on a 1-4 scale), and cue (included as a categorical variable). 

\emph{Motion Direction:} Linear regressions were performed relating peak displacement in x, y, z, yaw, pitch, and roll to the set of trials, subject (as a random variable), and cue (as a categorical variable). 
A multi-way ANOVA was performed comparing peak x, y, z, yaw, pitch, and roll displacements for each cue, followed by multiple comparison tests with the Bonferroni correction. 

\emph{Discrimination:} The discrimination task is analyzed by percent correct responses. A multi-way ANOVA was performed on the number of repeats by cue and by subject (as a random variable).

\section{Results}
\subsection{Motion Direction}
\begin{figure}[t]
  \centering
  \includegraphics[width=3.4in]{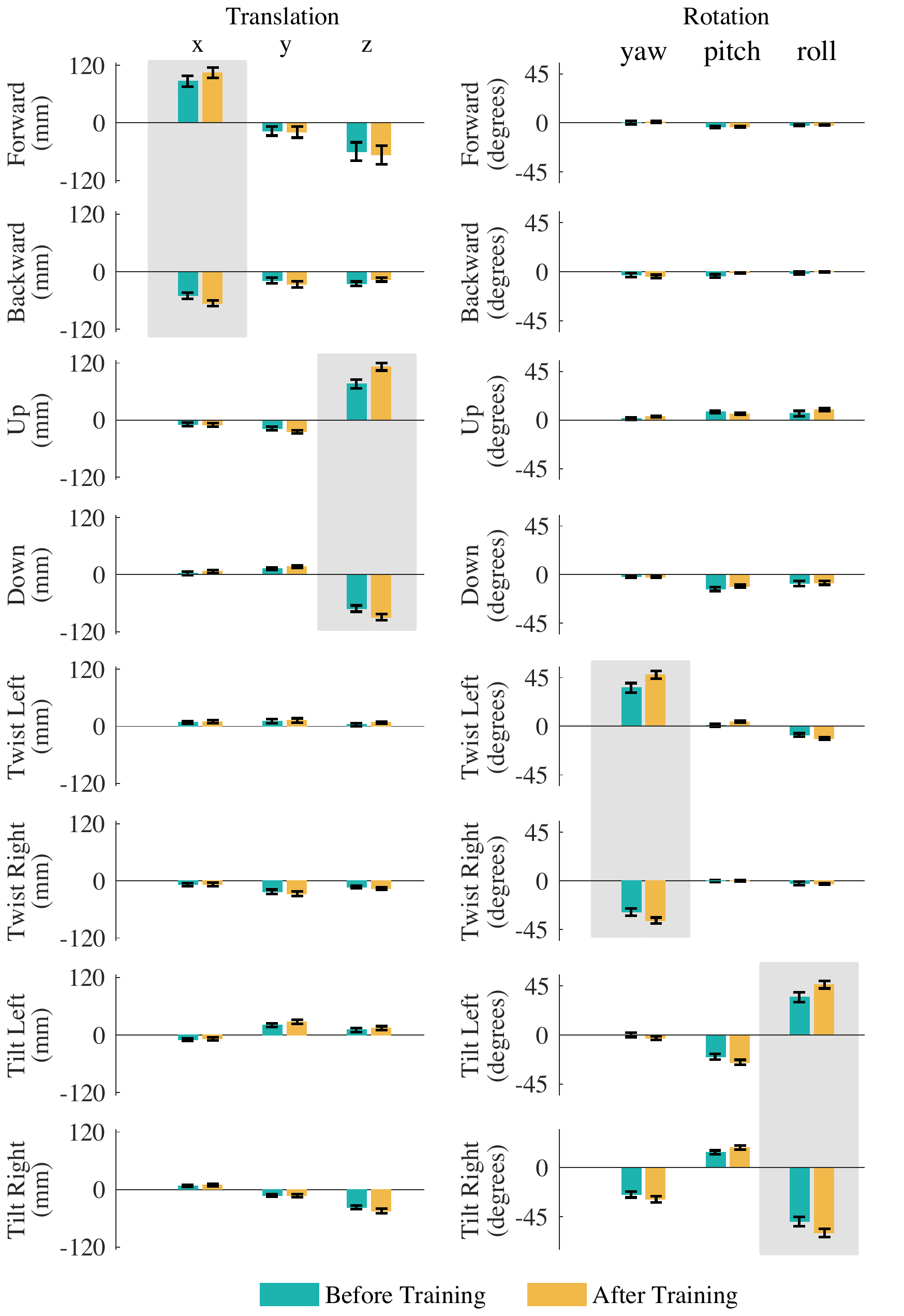}
  \caption{Mean peak displacements in Part 1 and Part 3 separated by cue direction. The intended motion direction is highlighted for each cue. Error bars show 95\% confidence intervals for the mean (n=200).}
  \label{fig:barplot}
\end{figure}

In both sets of movement experiments, users translated or rotated their hands primarily in the expected direction for each cue. Fig.~\ref{fig:allMovements} shows mean trajectories for each user separated by cue and DOF for all trials. Fig.~\ref{fig:barplot} shows the mean peak x, y, and z translation and yaw, pitch, and roll rotation separated by cue and set. 
Whether a trial was in the first or second set had no significant effect on the peak motion magnitude or direction.
Comparisons of peak x, y, z, yaw, pitch, and roll displacements showed that the two relevant cues had the largest effects, the two cues' effects opposed each other, and they were significantly different from all other cues' effects ($p<0.01$). Although pitch displacement was not targeted by any cue, there was a significant effect from tilt right and tilt left cues. For y displacement, no cues had a significant effect.

\subsection{Delay Before Motion}
\begin{figure}
  \centering
  \includegraphics[width=3.3in]{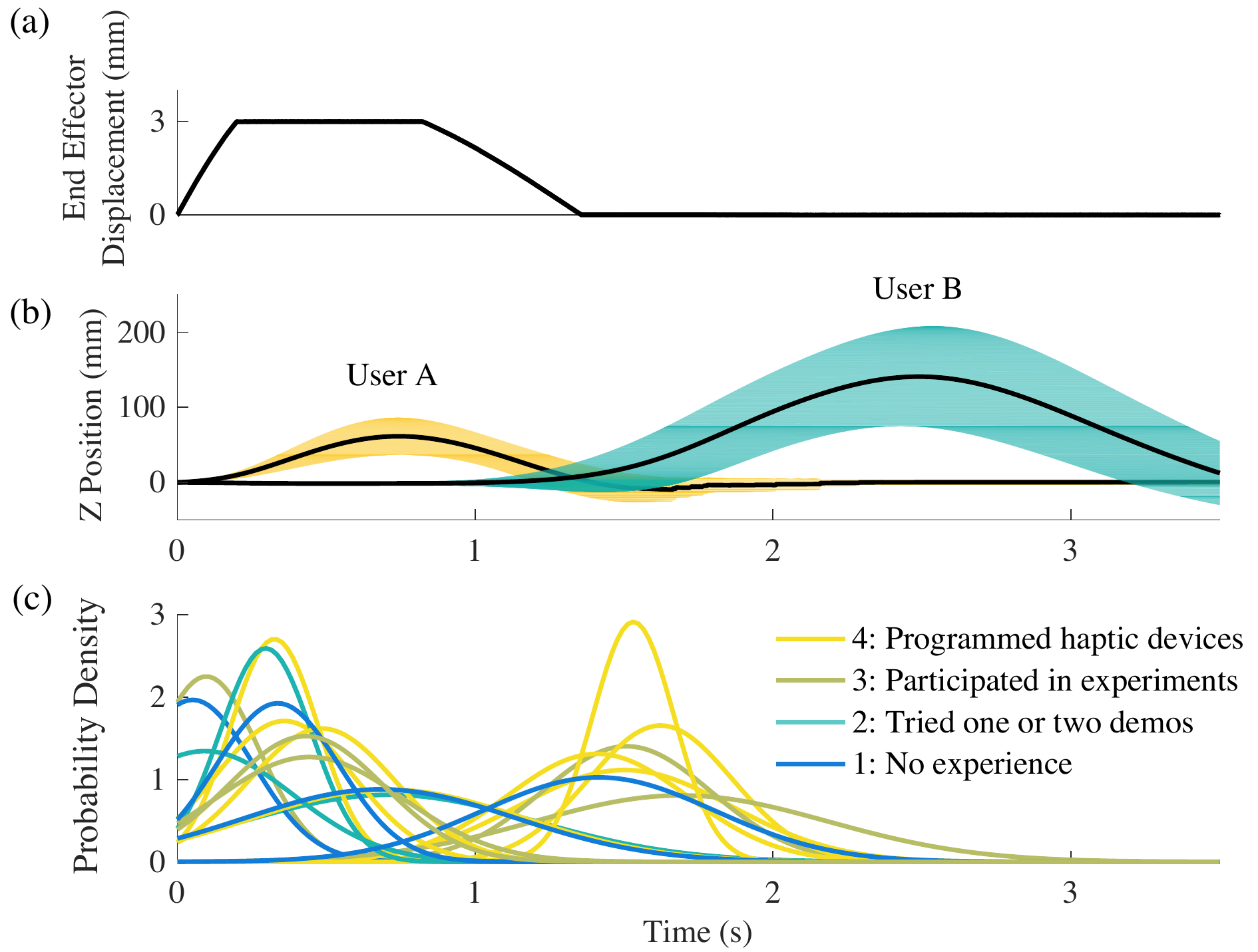}
  \caption{(a) Trajectory of end-effector during an \emph{up} cue. For all cues, the end-effectors followed this trajectory moving either horizontally or vertically. (b) Mean and standard deviations of two example users with different delay times. (c) Distribution of delay time for each subject. Colors indicate the subject's level of experience with haptic devices.}  
  \label{fig:delaydistributions}
\end{figure}
For all users, there was some delay between the start of a haptic cue and the onset of motion. Rotation cues had significantly longer delay times than translation cues (0.119 s lower on average, $p<0.01$). Set number did not have a significant effect on delay ($p>0.05$). The delay times have a bi-modal distribution with peaks at 0.33 s and 1.56 s. We will call these two groups \emph{fast responders} and \emph{slow responders}. Fig.~\ref{fig:delaydistributions}c shows the distribution of delay for each subject.
 Fig.~\ref{fig:delaydistributions}b shows example mean and standard deviation trajectories for a user from each of the distributions. Slower responses was also correlated with higher variance in peak displacement magnitude ($p<0.05$). Clustering by variance of peak magnitude and delay time reveals two distinct groups: 13 fast responders and 7 slow responders. More experience with haptic devices was correlated with a slightly lower delay time (-0.041 s/level, $p<0.001$). 

\subsection{Forced Choice Experiment}
Users were able to discriminate the eight cue directions over 93.3\% of the time in the discrimination task. Table~\ref{table:Confusion} shows the percent correct and mistaken for each cue direction. 
On average, users rated the difficulty of the device as 2.2 on a scale out of 4, corresponding to \emph{easy}. Several users reported that the rotation cues were more salient to them than the translation cues. 
There was no significant difference in the number of repeated cues between directions ($p>0.05$). 

\newcolumntype{R}[2]{%
    >{\adjustbox{angle=#1,lap=\width-(#2)}\bgroup}%
    l%
    <{\egroup}%
}
\newcommand*\rot{\multicolumn{1}{R{45}{1em}}}
\begin{table}
\caption{Confusion Matrix for Part 2: Forced Choice Experiment}
{\centering
\begin{tabular}[b]{ >{\centering}m{1.4cm}| >{\centering}m{0.4cm} >{\centering}m{0.4cm} >{\centering}m{0.4cm} >{\centering}m{0.4cm} >{\centering}m{0.4cm} >{\centering}m{0.4cm} >{\centering}m{0.4cm} m{0.4cm} }
&\multicolumn{8}{c}{\bfseries User Responses}\\
\bfseries Correct Direction &\rot{Forward} &\rot{Backward} &\rot{Up} &\rot{Down} &\rot{Twist Left} &\rot{Twist Right} &\rot{Tilt Left} & \rot{Tilt Right} \\ \hline
Forward & 96.7 \cellcolor[gray]{0.6} & 1.7 \cellcolor[gray]{0.95} & 1.7 \cellcolor[gray]{0.95} & 0.0 \cellcolor[gray]{1} & 0.0 \cellcolor[gray]{1} & 0.0 \cellcolor[gray]{1} & 0.0 \cellcolor[gray]{1} & 0.0 \cellcolor[gray]{1}\\ 
Backward & 1.7 \cellcolor[gray]{0.95} & 95.0 \cellcolor[gray]{0.65} & 0.0 \cellcolor[gray]{1} & 3.3 \cellcolor[gray]{.88} & 0.0 \cellcolor[gray]{1} & 0.0 \cellcolor[gray]{1} & 0.0 \cellcolor[gray]{1} & 0.0 \cellcolor[gray]{1}\\ 
Up & 1.7 \cellcolor[gray]{0.95} & 0.0 \cellcolor[gray]{1} & 93.3 \cellcolor[gray]{0.7} & 0.0 \cellcolor[gray]{1} & 1.7 \cellcolor[gray]{0.95} & 0.0 \cellcolor[gray]{1} & 0.0 \cellcolor[gray]{1} & 3.3 \cellcolor[gray]{0.88}\\ 
Down & 1.7 \cellcolor[gray]{0.95} & 1.7 \cellcolor[gray]{0.95} & 3.3 \cellcolor[gray]{0.88} & 91.7 \cellcolor[gray]{0.75} & 0.0 \cellcolor[gray]{1} & 0.0 \cellcolor[gray]{1} & 1.7 \cellcolor[gray]{0.95}  & 0.0 \cellcolor[gray]{1} \\ 
Twist Left & 0.0 \cellcolor[gray]{1} & 1.7 \cellcolor[gray]{0.95} & 0.0 \cellcolor[gray]{1} & 0.0 \cellcolor[gray]{1} & 93.3 \cellcolor[gray]{0.7} & 3.3 \cellcolor[gray]{0.88} & 1.7 \cellcolor[gray]{0.95} & 0.0 \cellcolor[gray]{1}\\ 
Twist Right & 0.0 \cellcolor[gray]{1} & 0.0 \cellcolor[gray]{1} & 0.0 \cellcolor[gray]{1} & 0.0 \cellcolor[gray]{1} & 1.7 \cellcolor[gray]{0.95} & 96.7 \cellcolor[gray]{0.6} & 0.0 \cellcolor[gray]{1} & 1.7 \cellcolor[gray]{0.95}\\ 
Tilt Left & 0.0 \cellcolor[gray]{1} & 0.0 \cellcolor[gray]{1} & 0.0 \cellcolor[gray]{1} & 1.7 \cellcolor[gray]{0.95} & 0.0 \cellcolor[gray]{1} & 0.0 \cellcolor[gray]{1} & 96.7 \cellcolor[gray]{0.6} & 1.7 \cellcolor[gray]{0.95}\\ 
Tilt Right & 0.0 \cellcolor[gray]{1} & 0.0 \cellcolor[gray]{1} & 0.0 \cellcolor[gray]{1} & 0.0 \cellcolor[gray]{1} & 0.0 \cellcolor[gray]{1} & 3.3 \cellcolor[gray]{0.88} & 1.7 \cellcolor[gray]{0.95} & 95.0 \cellcolor[gray]{0.65} \\ 

\end{tabular}
\label{table:Confusion}}
\footnotesize{\\**Cells are shaded corresponding to percentage of user responses.}
\end{table}

\section{Discussion}

There are several key findings from the user study. First, our handheld kinesthetic device can provide clear guidance in 4-DOF. Second, the haptic cues are intuitive enough that there is very little delay before users begin moving. Finally, there is variability in the ways different users respond to holdable guidance cues.   

\subsection{Differentiable Cues}
All cues were successfully identified during the forced choice experiment in more than 90\% of the trials. Even more encouraging is that the users' motion responding to each cue in Part 1 and Part 3 was primarily in the intended direction (Fig.~\ref{fig:barplot}). 
Although the actuation was always the same magnitude, the magnitude of the response motion varies by direction and user. Human force perception is anisotropic~\cite{VanBeek2013}. For ungrounded haptic devices, these anisotropies may be even more extreme, and they may need to be characterized per device or per user. Users in our study moved farther forward than backward. They also tilted farther right than left. We did not study different cue magnitudes or speeds in this initial study, but we believe users could be cued to move farther or for more time by adjusting the cue magnitude or length. This seems especially likely for the fast responders, who started and stopped their movements in time with the end-effectors' motion and had more consistent peak displacements (Fig.~\ref{fig:delaydistributions}(b)).  

Responses to some cues had biases toward uncued directions. For four users, there was notable coupling between forward x and downward z motion in response to forward cues (Fig.~\ref{fig:allMovements}). There was also a small amount of pitch when responding to roll cues. It is possible that by adjusting the end-effector actuation, pure roll could be isolated. Alternatively, this suggests it might be possible to provide some guidance for pitch. We will consider these ideas in future work. 

\subsection{Intuitive Direction Guidance}
Even before the cues were explained in Part 2, users moved in the intended directions. (Fig.~\ref{fig:barplot}, before training). Training in Part 2 did not significantly reduce delay or change users' motions. 
This implies the ungrounded guidance cues are intuitive and could be applied without training in virtual environments or teleoperation tasks.
Most users responded quickly to each haptic cue. Fast responders moved their hands while the haptic cue was still being applied, or shortly after it. 
The mean delay for fast responders was 0.33s, which is on the order of the human reaction time to touch~\cite{Lele1954}, implying that very little interpretation was needed. Slow responders, with a mean delay of 1.56 s, waited for the end-effectors to return to the center point before moving their hand. They felt the cue, interpreted the cue, then responded. Because experience with haptic devices slightly reduced delay (Fig.~\ref{fig:delaydistributions}(c)), it may be possible that additional practice could transform slow responders into fast responders. This effect emphasizes the importance of including novice users in all haptics experiments. 

Although we did not include non-cardinal directions in this preliminary study, users did not need to know the cues to be able to follow them. So, we are optimistic that a continuous range of directions could also be applied with good performance. This seems especially likely for the fast responders who moved as if they were being pulled by an external force. We will study a range of magnitudes, speeds, and directions in future research.

\subsection{Variability Between Users}
Holdable and wearable devices cannot enforce user motion like grounded devices can. Some differences are expected between users, as in Fig.~\ref{fig:allMovements}. 
Psychometric tests on perception in different directions might clarify these differences and should be explored. However, users' responses to haptic guidance are affected by perception, hand and wrist kinematics, and some user interpretation, making it difficult to fully explain their motion. 
Alternatively, it may be helpful to adapt guidance cues to each user. As interactions are recorded, the system could develop a model of the users' responses. 
The directions of the cues could be rotated or the magnitudes scaled to balance each users' trends.

\section{Conclusion}
The holdable haptic device presented uses pantograph five-bar-linkages to displace the index finger and thumb. It provides intuitive guidance in 4-DOF without external grounding. The haptic cues did not require training to produce motion in each direction tested, and the cues could be discriminated easily. Application areas include teleoperation, virtual or augmented reality, and training simulations for manipulation tasks like surgery. 


\addtolength{\textheight}{-12cm}   







\bibliography{WHC2019.bib}{}
\bibliographystyle{IEEEtran}

\end{document}